# Force-Compliance MPC and Robot-User CBFs for Interactive Navigation and User-Robot Safety in Hexapod Guide Robots

Zehua Fan, Feng Gao*, Zhijun Chen, Yunpeng Yin, Limin Yang, Qingxing Xi, En Yang, Xuefeng Luo

*Abstract*—Guiding the visually impaired in complex environments requires real-time two-way interaction and safety assurance. We propose a Force-Compliance Model Predictive Control (FC-MPC) and Robot-User Control Barrier Functions (CBFs) for force-compliant navigation and obstacle avoidance in Hexapod guide robots. FC-MPC enables two-way interaction by estimating user-applied forces and moments using the robot's dynamic model and the recursive least squares (RLS) method, and then adjusting the robot's movements accordingly, while Robot-User CBFs ensure the safety of both the user and the robot by handling static and dynamic obstacles, and employ weighted slack variables to overcome feasibility issues in complex dynamic environments. We also adopt an Eight-Way Connected DBSCAN method for obstacle clustering, reducing computational complexity from $O(n^2)$ to approximately $O(n)$, enabling real-time local perception on resource-limited on-board robot computers. Obstacles are modeled using Minimum Bounding Ellipses (MBEs), and their trajectories are predicted through Kalman filtering. Implemented on the HexGuide robot, the system seamlessly integrates force compliance, autonomous navigation, and obstacle avoidance. Experimental results demonstrate the system's ability to adapt to user force commands while guaranteeing user and robot safety simultaneously during navigation in complex environments.

*Note to Practitioners*—Guiding visually impaired individuals in complex environments remains a significant challenge, as traditional methods like guide dogs are limited by high costs and scalability issues. To address this, we propose a Hexapod guide robot framework that integrates Force-Compliance Model Predictive Control (FC-MPC) and Robot-User Control Barrier Functions (CBFs) to ensure real-time two-way interaction and dual safety for both the user and the robot. The framework supports diverse interaction modes by allowing parameter adjustments to combine functionalities such as force compliance, global autonomous navigation, and obstacle avoidance. An efficient clustering algorithm is adopted in the local environment perception process, which enables real-time deployment on resource-constrained devices of guide robot. While effective in experiments, future work will focus on map-less navigation to improve performance in previous unseen environments.

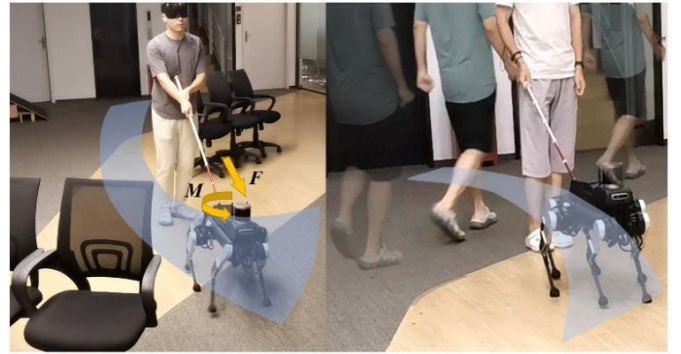

Fig. 1. Two-way force interaction control locomotion with obstacle avoidance (left) and dynamic obstacle avoidance in force-compliant autonomous navigation under the FC-MPC and Robot-User CBF constraints (right) for the HexGuide guide robot.

## I. INTRODUCTION

TRAVELING outdoors has always been a big challenge for the visually impaired people. Although guide dogs offer a solution, their training is time-intensive, expensive, and limited in output, insufficient to meet the needs of a larger blind population. Recently, electronic guide devices, especially legged robots, have gained attention due to their obstacle-crossing capabilities [1] .

When guiding visually impaired individuals, guide robot must follow the user's intentions at a global level, executing their movement commands. At a local level, the robot must selectively override the user's intentions in order to safely avoid obstacles that the blind user cannot perceive. This necessitates the robot's ability to adapt to the surrounding environment in real time (environmental perception) while also comprehending the user's commands (user command interpretation).

In terms of environmental perception, prior research has predominantly employed LiDAR [2, 3] or cameras [4] to capture point cloud data for localization and obstacle detection, followed by clustering process and obstacle parameterization [5]. Regarding user command interpretation, as visually impaired individuals cannot rely on visually-dependent interactions, voice interaction [6] and force-based interaction [7] are the main approaches. Given the stringent real-time requirements during robot navigation, voice interaction is less suitable. This makes force-based interaction, where the user applies movement commands directly to the robot, a more appropriate solution.

We aim for the guide robot to balance user-applied force commands with the surrounding environmental obstacles to make appropriate navigation decisions. Furthermore, during obstacle avoidance, the safety considerations should not only apply to the robot but also to the user. To achieve this, in this work, we designed a novel control framework for guide dog. We employ a force estimation method based on the robot's dynamic model and the recursive least squares (RLS) method to capture the user's force interaction commands. We then designed a local planner called **F**orce-**C**ompliance **M**odel **P**redictive **C**ontroller **(FC-MPC)** to integrate these commands into local planning. Additionally, we model and construct the Control Barrier Functions of both the robot and the blind user, denoted as **Robot-User CBFs**. These CBFs are incorporated into FC-MPC as soft constraints using weighted slack variables, enabling the planner to jointly consider the robot-user joint safety, while prioritizing the user's safety in the presence of potential conflicts. To implement this system on the computational resource-constrained HexGuide robot, we also proposed to use an Eight-Way Connected DBSCAN clustering method, which reduces the algorithmic time complexity from $O(n^2)$ in traditional DBSCAN clustering methods to approximately $O(n)$. This ensures that obstacle perception can operate at a higher frequency, maintaining the real-time performance of the local planner on resource-constrained onboard computers. Through the implementation of our proposed framework, we achieved a seamless integration of force compliance, local static and dynamic obstacle avoidance, and global navigation in the guiding process. Shortcuts of typical experiments are shown in Fig. 1;

### A. Related Work

#### (1) User-Robot Interaction

Guide robots come in various forms: wheeled [8], quadruped [3, 9], and hexapod robots [10]. Legged robots are better suited for complex terrains, while hexapods offer greater stability, enhancing user comfort. Traditionally, guide robots connect with the user through rigid rods [4, 8, 10], robotic arms [11], or flexible tethers [3, 9, 12]. Some studies focus on optimizing tether forces for user comfort [3]. In [11], the authors introduced an adaptive pulling planner to guide visually impaired individuals back onto the correct path when they deviate from the intended route. However, these methods primarily offer **one-way** guidance.

While using the guide dog, the blind users prefer controlling the speed and direction of guide dog on their own, rather than entirely depending on the dog. They typically determine how to move by means of smartphone navigation apps, auditory cues, partial light perception, or guidance from others. To support this preference, it becomes essential to introduce **two-way force interaction** into the navigation algorithm, allowing the user's physical input to influence the robot's motion planning. The use of rigid connections, such as guide canes, is crucial in this context, as flexible tethers cannot effectively transmit forces—particularly pushing forces—back to the robot. Therefore, we choose to use a rigid guide cane to connect the user and implement dynamic model-based external force estimation and compliance. We integrate user commands into navigation using an FC-MPC local planner, enabling seamless two-way physical interaction and adaptive motion control.

In addition to hardware modalities, prior works have also modeled the human–robot system using coupled kinematic or dynamic formulations, depending on the assumed sensing and actuation capabilities. A seminal formulation in this direction is the Leader–Follower (L–F) framework proposed by Panagou and Kumar [13], where the robot leads and another agent follows under relative pose or visibility constraints. In this setup, the follower estimates the leader's position using onboard vision and adjusts its own motion accordingly. Later extensions adapted this paradigm to human–robot systems via physical force transmission, such as hybrid leash-tension modeling [9], tractor–trailer kinematics [14], or comfort-optimized force-based planning [3], where the robot modulates its motion to ensure smooth and predictable user responses. These formulations provide valuable theoretical foundations, but most assume that the follower either possesses visual feedback or passively follows the leader's intended path under unidirectional control.

In contrast, our setup considers the human user as an active participant who, though lacking visual feedback, can apply physical forces at any time to influence the robot's motion. We introduce an external-force-aware, compliance-based MPC planner that enables shared control, where either the robot or the user may assume initiative. We use the robot and the blind person's states (i.e., robot-user co-state) together to the control system. In addition, to ensure safety, robot–user coupled CBFs are applied. This formulation extends the L–F framework by supporting bidirectional physical interaction, which is essential in assistive guidance tasks with human intention involved.

#### (2) Environmental Perception and Local Planning

Safe navigation requires effective global path planning [15-19], as well as the local perception and obstacle model, followed by local planning to avoid the static and dynamic obstacles.

Sensors like LiDAR, depth cameras or event cameras gather point cloud data of obstacles. Then the data points are clustered by clustering algorithm such as DBSCAN [2, 20, 21] and k-nearest neighbors algorithm [22]. When using DBSCAN on computational resource-constrained devices on robots, it is necessary to make certain improvements to the traditional algorithms to enhance real-time performance [23]. Inspired by [21] which achieves low computation time cost on computational resource-constrained quadrotor, we adopted and modify an Eight-Connected DBSCAN clustering algorithm for guide dog. This method outperforms the traditional DBSCAN algorithm by a large margin.

After clustering process, the clustered obstacles are modeled using methods like minimum bounding ellipses (MBEs) [2] and minimum volume ellipsoids (MVEs) [24]. For guide dog, we adopt MBEs since it is a planar problem.

Prior research explores various local planning methods, such as DWA [25], artificial potential fields [26],

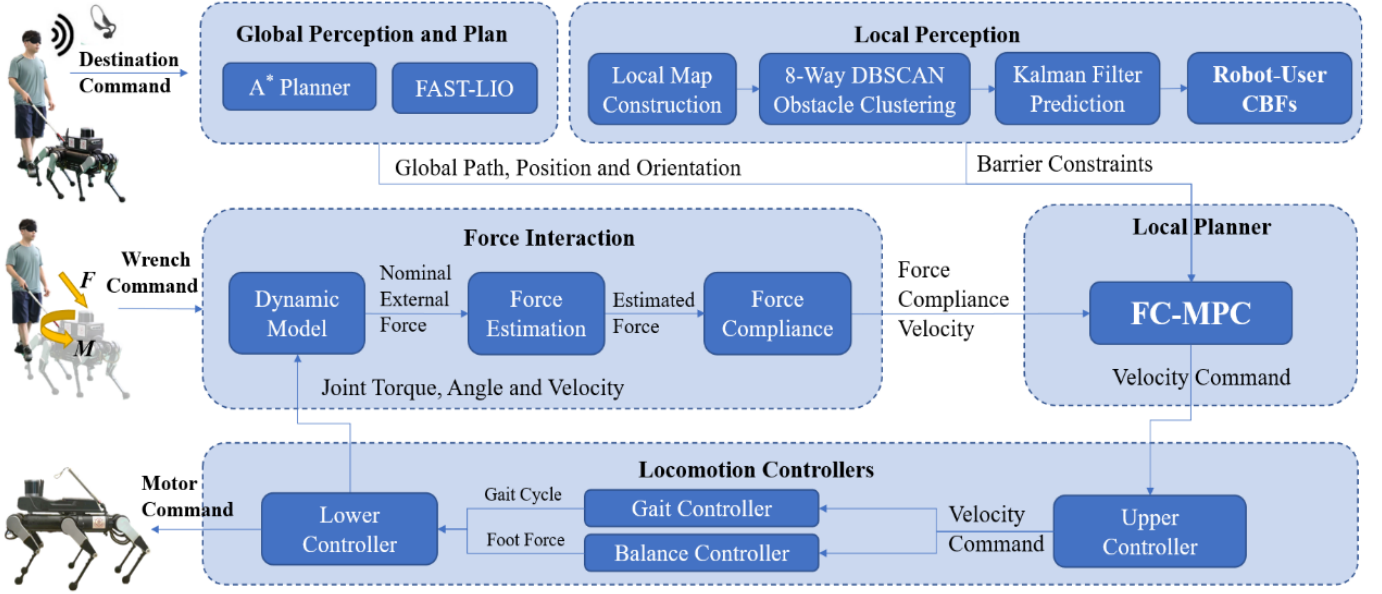

Fig. 2.    Our proposed guide robot control framework.

reinforcement learning [27-29], space-time navigation controller [30], LSTM Based Predictive Model [31] and MPC with constraints for obstacle avoidance [2, 32-34]. However, when these local planning methods are applied to guide robots without modification, only the safety of the robot itself is considered while the safety of the blind user is ignored. In [3, 4, 35], the existence of the blind user is taken into consideration and good performance is achieved. However, they do not clearly give clear mathematical safety constraints for obstacle avoidance of the robot and user. This makes it difficult for them to theoretically guarantee that the system is always within the safe area or returns to a safe state within a limited number of time steps.

Recently, Control Barrier Function (CBF) has attracted the attention of many researchers [36, 37]. CBFs have been successfully applied to both static [38] and dynamic obstacles [2]. Notably, [34] integrated CBFs as constraints into MPC, creating an efficient framework for real-time obstacle avoidance. In prior works, CBFs are often enforced as hard constraints to guarantee safety [2, 34, 38, 39]. However, in complex environments with numerous dynamic obstacles, the robot may unexpectedly fall outside their designated safety sets due to obstacle high speed movements, leading to constraint violations that render the local planner temporarily infeasible and significantly impair navigation performance. To address this, slack variables have been introduced to relax the hard constraints, enabling CBFs to function as soft constraints within the optimization [40]. When the local planner accounts for both the robot and the user, their co-state must also be incorporated into the CBF-based safety framework [39]. In such co-state-aware planning for complex dynamic obstacles, **four critical failure modes** may arise: (1) the robot's state may fall outside its safety set, (2) the user's state may fall outside its safety set, (3) both may do so simultaneously, or

(4) the robot and user safety sets may become disjoint—leaving no feasible region that guarantees mutual safety. Previous CBF frameworks were incapable of simultaneously handling all four failure scenarios described above.

Existing hard-constrained CBF frameworks (e.g., [2, 34, 38]) can typically handle Mode (1) (robot unsafe) in predictable cases, but often become infeasible under sudden constraint violations. Soft-constrained CBFs using slack variables (e.g., [40]) improve robustness for Mode (1), but rarely extend to user-state considerations. Works modeling robot-human co-states (e.g., [39]) can handle Modes (1) and (2), and implicitly adapt to Mode (3), but lack explicit safety prioritization or infeasibility recovery. While some prior works have explored aspects of failure mode handling, few have explicitly addressed Modes (3) and (4)—particularly under real-time co-state-aware optimization. Therefore, we incorporated weighted slack variables into co-state-based CBFs and explicitly prioritizing the user's safety under constraint conflicts, which simultaneously accounts for all four failure modes within a unified optimization framework.

We propose a Robot-User CBFs formulation with weighted slack variables, integrated as soft constraints into the FC-MPC. By explicitly prioritizing the safety of the user's state when conflicts arise, our method robustly maintains feasibility while ensuring safe and consistent navigation. To the best of our knowledge, this is the first time that the robot human co-state Control Barrier Function has been used in a guide robot.

### B. Contributions

The main contributions of this paper are as follows:

- We propose a novel local planner, Force-Compliance Model Predictive Control (FC-MPC), which estimates external forces using a dynamic model and recursive least squares (RLS), and integrates user-applied forces into local planning. This enables two-way interaction between the guide robot and the visually impaired user and is applicable to other robots requiring force compliance.

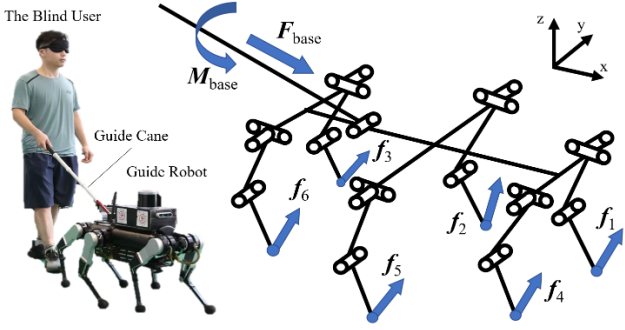

Fig. 3.    The guide robot and its force interaction diagram.

- We introduce Robot-User Control Barrier Functions (CBFs) into FC-MPC as novel soft constraints based on robot-human co-state. By modeling and predicting static and dynamic obstacles, this method ensures user and robot safety during obstacle avoidance. Weighted slack variables are used to resolve infeasibility arising from co-state-based CBFs in complex dynamic environments.

- By integrating FC-MPC with Robot-User CBFs, we developed a control framework that was successfully deployed on the HexGuide robot. Experimental results demonstrate that this framework seamlessly combines global autonomous navigation, compliance with the user's force commands, and safe avoidance of both static and dynamic obstacles, ensuring robot-user joint safety.

- We proposed to use an Eight-Way DBSCAN clustering method for obstacle perception, reducing the time complexity from O($n^2$) to approximately O($n$). This ensures that the proposed algorithm can be implemented on guide robots with limited computational resources.

## II. OVERVIEW OF THE FRAMEWORK

Our proposed guide robot control framework is illustrated in Fig. 2. The blind user provides destination commands via voice input, while mapping and localization are handled by the FAST-LIO [41]. Global path planning algorithm is the A* algorithm [42]. Locally, LiDAR point clouds are used to construct grid local map. For obstacle perception, we adopt a Eight-Way Connected DBSCAN Clustering Method, and obstacles are modeled using MBEs. We use Kalman Filter to predict the trajectory of dynamic obstacles. After that, both the robot and the user are incorporated into the Guide Control Barrier Functions (Robot-User CBFs) for building constraints of FC-MPC local planner. Force estimation is based on the robot's dynamic model and RLS method. According to the estimated external force, force compliance velocity is computed using an impulse-adaptive algorithm. The force-compliant velocity tracking goals are integrated into an MPC-based local planner, constrained by Robot-User CBFs, forming the FC-MPC. This enables real-time two-way force interaction while avoiding static and dynamic obstacles. The velocity commands from FC-MPC are sent to the robot's locomotion controller, which converts them into motor inputs.

Section II presents the overall framework of the guide robot algorithm. Section III.A describes the force interaction, including force estimation, compliance algorithm and the proposed FC-MPC local planner. Section III.B introduces the local perception process. The obstacle perception and the Robot-User CBFs are elaborated in detail. Section IV provides experimental validation of the system's performance.

## III. IMPLEMENTATION

### A. Force Interaction

To achieve two-way force interaction between the blind user and the robot, we estimate the external forces and moments acting on the robot using its dynamic model. These estimated forces are integrated into the MPC as compliant velocity terms, forming the Force Compliance Model Predictive Control.

#### (1) Dynamic Model-Based External Force Estimation

As is shown in Fig. 3, we model the guide robot as a six-degree-of-freedom floating base system, subjected to external forces applied by the user through the guide cane and the supporting forces on the robot's six feet. The dynamics equation is as follows:

$$\mathbf{M}(\boldsymbol{q})\ddot{\boldsymbol{q}} + \mathbf{C}(\boldsymbol{q},\dot{\boldsymbol{q}}) + \mathbf{G}(\boldsymbol{q}) = \boldsymbol{\tau} + \mathbf{J}^T \boldsymbol{\Lambda}, \qquad (1)$$

where $\mathbf{M}(\boldsymbol{q})$ is the generalized inertia matrix of the robot, $\mathbf{C}(\boldsymbol{q},\dot{\boldsymbol{q}})$ represents the Coriolis and centrifugal forces, and $\mathbf{G}(\boldsymbol{q})$ accounts for gravity. $\boldsymbol{q} = \begin{bmatrix} \boldsymbol{p}_{\text{base}} & \boldsymbol{\theta}_{\text{base}} & \boldsymbol{q}_{\text{motor}} \end{bmatrix}^T \in \mathbb{R}^{24}$ is defined as the generalized coordinates, including the position $\boldsymbol{p}_{\text{base}} \in \mathbb{R}^3$ and orientation $\boldsymbol{\theta}_{\text{base}} \in \mathbb{R}^3$ of the floating base, along with the joint angles of the six legs' 18 motors $\boldsymbol{q}_{\text{motor}} \in \mathbb{R}^{18}$. The generalized driving force is $\boldsymbol{\tau} = \begin{bmatrix} \mathbf{0}_{6\times1} & \boldsymbol{\tau}_{\text{motor}} \end{bmatrix} \in \mathbb{R}^{24}$. As the dynamic model of the guide robot is designed as a floating base model, the first six elements of $\boldsymbol{\tau}$ are zeros. The subsequent 18 elements correspond to the output torques $\boldsymbol{\tau}_{\text{motor}} \in \mathbb{R}^{18}$ of the 18 motors distributed across the six legs of the robot. $\boldsymbol{\Lambda} \in \mathbb{R}^{24}$ is the generalized external force vector comprising 24 elements in total. The detailed definition is:

$$\boldsymbol{\Lambda} = \begin{bmatrix} \boldsymbol{\Lambda}_{\text{base}} \\ \boldsymbol{\Lambda}_{\text{foot}} \end{bmatrix} = \begin{bmatrix} \boldsymbol{F}_{\text{base}} \\ \boldsymbol{M}_{\text{base}} \\ \boldsymbol{f}_1 \\ \vdots \\ \boldsymbol{f}_6 \end{bmatrix}. \qquad (2)$$

The first 6 elements represent the 6-dimensional wrench $\boldsymbol{\Lambda}_{\text{base}} = \begin{bmatrix} \boldsymbol{F}_{\text{base}} & \boldsymbol{M}_{\text{base}} \end{bmatrix}^T \in \mathbb{R}^6$ applied to the floating base, where $\boldsymbol{F}_{\text{base}} \in \mathbb{R}^3$ and $\boldsymbol{M}_{\text{base}} \in \mathbb{R}^3$ are the force and moments applied by the blind user to the guide robot, which we aim to estimate. The remaining 18 elements of the $\boldsymbol{\Lambda}$ correspond to the 3-dimensional forces acting on each of the six spherical foot tips of the robot, denoted as $\boldsymbol{\Lambda}_{\text{foot}} \in \mathbb{R}^{18}$.

$\mathbf{J}^T$ represents the generalized Jacobian matrix, which is defined as follows:

$$\mathbf{J} = \begin{bmatrix} \mathbf{I}_{6\times6} & \mathbf{0}_{6\times18} \\ \mathbf{J}_{\text{bl}} & \mathbf{J}_{\text{foot}} \end{bmatrix}, \tag{3}$$

where $\mathbf{I}_{6\times6}$ is the identity matrix, $\mathbf{J}_{\text{foot}} \in \mathbb{R}^{18\times18}$ is the Jacobian matrix of the legs with 3 degree of freedoms, and $\mathbf{J}_{\text{bl}} \in \mathbb{R}^{18\times6}$ is the relationship matrix between the base and the leg.

The dynamic characteristics of the robot are denoted as $\boldsymbol{\tau}_{\text{dyn}}$, which are given by:

$$\boldsymbol{\tau}_{\text{dyn}} = \mathbf{M}(\boldsymbol{q})\ddot{\boldsymbol{q}} + \mathbf{C}(\boldsymbol{q},\dot{\boldsymbol{q}}) + \mathbf{G}(\boldsymbol{q}). \tag{4}$$

The onboard IMU provides measurements of the robot body's linear acceleration $\dot{\vec{p}}_{\text{base}}$, angular velocity $\boldsymbol{\omega}$, and angular acceleration $\dot{\boldsymbol{\omega}}$. Joint torques $\boldsymbol{\tau}_{\text{motor}}$ can be estimated based on the current measured from the joint actuator. Joint position $\boldsymbol{q}_{\text{motor}}$ and joint velocity $\dot{\boldsymbol{q}}_{\text{motor}}$ can be obtained from the leg-mounted joint sensors. Using numerical differentiation or a Kalman-based state observer, the joint angular acceleration $\ddot{\boldsymbol{q}}_{\text{motor}}$, as well as the robot's position $\boldsymbol{p}_{\text{base}}$, orientation $\boldsymbol{\theta}_{\text{base}}$, and linear velocity $\dot{\boldsymbol{p}}_{\text{base}}$, can be estimated. Based on these quantities, the nominal external force can then be computed, denoted as $\boldsymbol{\tau}_{\text{obs}} = \boldsymbol{\tau}_{\text{dyn}} - \boldsymbol{\tau}$. We use the recursive least squares (RLS) method for external force estimation, introducing a forgetting factor $\alpha \in (0,1)$ allows the robot to "forget" past information and focus more on the most recent force data. Let the estimation error caused by the system and measurement be denoted as $\tilde{\boldsymbol{e}} = \mathbf{J}^T \boldsymbol{\Lambda} - \boldsymbol{\tau}_{\text{obs}}$. By minimizing the error, the estimated external force at the current time, $\hat{\boldsymbol{\Lambda}}_{\text{base}}(t_N)$, is obtained as follows:

$$\hat{\boldsymbol{\Lambda}}_{\text{base}}(t_N) = \underset{\boldsymbol{\Lambda}(t_n)}{\arg\min}\left(\sum_{n=1}^{N} \alpha^{N-n}\tilde{\boldsymbol{e}}(t_n)_{\mathbf{D}}^2\right), \tag{5}$$

where $N = t/\Delta t$ is the integer-rounded value of the current time $t$, and $\Delta t$ is the time step length determined by the estimation frequency. $\tilde{\boldsymbol{e}}(t_n)_{\mathbf{D}}^2$ indicates the square norm of $\tilde{\boldsymbol{e}}(t_n)$ with weight matrix $\mathbf{D}$.

By analogy with the Kalman filter method, the covariance matrix and gain matrix can be calculated using the following formulas:

$$\mathbf{P}_n = \frac{1}{\alpha}\left(\mathbf{I}_{24\times24} - \mathbf{K}_n\mathbf{J}_n^T\right)\mathbf{P}_{n-1}, \tag{6}$$

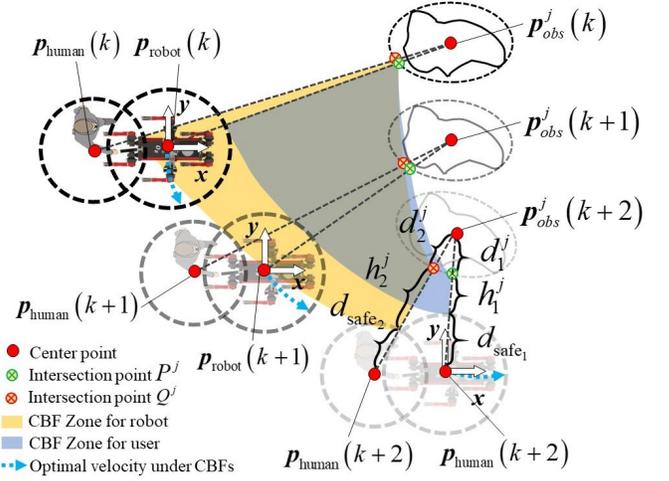

Fig. 4. Schematic diagram showing the optimal velocity output of HexGuide by FC-MPC under the CBF constraints, leading the robot and the blind user to avoid dynamic obstacles modeled using the MBE method.

$$\mathbf{K}_n = \mathbf{P}_{n-1}\left(\mathbf{J}_n^T\right)^T\left(\alpha\mathbf{I}_{24\times24} + \mathbf{J}_n^T\mathbf{P}_{n-1}\left(\mathbf{J}_n^T\right)^T\right)^{-1}, \tag{7}$$

where $\mathbf{P}_n$, $\mathbf{K}_n$, and $\mathbf{J}_n$ represent the error covariance matrix, Kalman gain, and generalized Jacobian matrix at time $t_n$, respectively. The update formula for the external force estimation is as follows:

$$\hat{\boldsymbol{\Lambda}}_n = \hat{\boldsymbol{\Lambda}}_{n-1} + \mathbf{K}_n\left(\boldsymbol{\tau}_{obs_n} - \mathbf{J}_n^T\mathbf{P}_{n-1}\left(\mathbf{J}_n^T\right)^T\right)^{-1}, \tag{8}$$

where $n$ as a subscript represents the value of the variables at time $t_n$. The first 6 rows of $\hat{\boldsymbol{\Lambda}}_n$ are extracted as the estimated external force and moments vector acting on the guide robot's base:

$$\begin{bmatrix} \hat{\boldsymbol{F}}_{\text{base}_n} \\ \hat{\boldsymbol{M}}_{\text{base}_n} \end{bmatrix} = \hat{\boldsymbol{\Lambda}}_{\text{base}_n} = \left[\hat{\boldsymbol{\Lambda}}_n\right]_{1-6}. \tag{9}$$

The guide robot needs to comply with the estimated external forces and moments applied by the blind user. We are primarily interested in the forces along the x-axis and y-axis, as well as the moments around the z-axis. First, the impulse in the corresponding directions is obtained through integration:

$$\boldsymbol{L}_N = \sum_{n=N-N_r+1}^{N} \gamma^{N-n}\begin{bmatrix} \left[\hat{\boldsymbol{F}}_{\text{base}_n}\right]_x \\ \left[\hat{\boldsymbol{F}}_{\text{base}_n}\right]_y \\ \left[\hat{\boldsymbol{M}}_{\text{base}_n}\right]_z \end{bmatrix}\Delta t, \tag{10}$$

where $\boldsymbol{L}_N$ represents the accumulated impulse of the external force from the starting discrete time $N - N_r + 1$ to the current time $N$, where $N_r$ is the number of the most recent time steps we are interested in for accumulating impulse. The forgetting factor $0 < \gamma < 1$ ensures that the robot's force compliance focuses more on the most recent force-induced impulse, while also guaranteeing computational convergence.

Furthermore, to comply with the external force applied by the user, the current force compliance velocity $\boldsymbol{v}_{\text{tgt}_N}$ can be obtained using the following equation:

$$v_{\text{tgt}_N} = v_{\text{tgt}_0} + \mathbf{W}^{-1} L_N, \tag{11}$$

where $v_{\text{tgt}_0}$ is the velocity of the guide dog at time step $N - N_r + 1$. The matrix $\mathbf{W} = \text{diag}\left(m, m, J_z\right)$ is the inertia matrix. $m$ is the mass of the guide robot, and $J_z$ is the moment of inertia corresponding to the rotational direction around the z-axis.

From the method mentioned above, we know that hexapod robots, due to their structural and dynamic characteristics, can estimate external forces and moments solely based on motor current feedback without relying on additional force sensors, making them both cost-effective and reliable. This is why we chose hexapod robots as guide robots in this study.

*(2) Force Compliance Model Predictive Control*

To use the MPC method as a local planning strategy, the discrete state-space equations of the guide robot must first be established. The guide robot has omnidirectional movement capabilities in three directions, and its kinematic equations are as follows:

$$
\begin{aligned}
x_{k+1} &= \mathbf{F} x_k + \mathbf{H}_k u_k \Delta t \\
&= x_k + \begin{bmatrix} \cos\theta_k & -\sin\theta_k & 0 \\ \sin\theta_k & \cos\theta_k & 0 \\ 0 & 0 & 1 \end{bmatrix} u_k \Delta t,
\end{aligned} \tag{12}
$$

where the state vector $x_k = \begin{bmatrix} x_k & y_k & \theta_k \end{bmatrix}^T$ represents the position and orientation of the robot in the global coordinate system, and the input vector $u_k = \begin{bmatrix} v_{x_k} & v_{y_k} & \omega_{z_k} \end{bmatrix}^T$ denotes the velocity of the robot in the forward, lateral, and yaw rotational directions. The robot coordinate system is illustrated in Fig. 4.

To enable navigation and obstacle avoidance while complying with the user's force commands, we integrate the current force compliance velocity $v_{\text{tgt}_N}$ mentioned in Section III.A.(1) into the MPC optimization function, forming the **force compliance term** $v(u_k)$. The Force Compliance MPC optimization problem is formulated as follows:

$$\min_{u_{0:Z-1}} p(x_Z) + \sum_{k=0}^{Z-1} q(x_k) + \sum_{k=0}^{Z-1} v(u_k) \tag{13a}$$

s.t. $x_{k+1} = \mathbf{F} x_k + \mathbf{H} u_k \Delta t, \ k = 0, 1, ..., Z-1$ (13b)

$x_k \in \chi, u_k \in U, k = 0, 1, ..., Z-1$ (13c)

$x_0 = x_t, x_Z \in X_f$ (13d)

$$\Delta h_i(X_k) \ge -\beta h_i(X_k), \begin{cases} \beta \in (0,1] \\ k = 0, 1, ..., Z-1 \\ i = 1, 2 \\ j = 1, \ 2, \ ..., m \end{cases} \tag{13e}$$

, where the subscript $t+k|t$ is denoted as $k$, and Z represents the time steps in the model prediction. In the formula (13a) $v(u_k) := \left\| u_k - \mu^k v_{\text{tgt}_N} \right\|_{\mathbf{R}} + \left\| u_k - u_{k-1} \right\|_{\mathbf{S}}$ represents the cost of force compliance at future time step $k$, where $\left\| * \right\|_{\mathbf{R}}$ and $\left\| * \right\|_{\mathbf{S}}$ denotes a norm with the weight matrix $\mathbf{R}$ and $\mathbf{S}$, respectively. $v_{\text{tgt}_N}$ is the most recent force compliance velocity command calculated at current time (step $k$=0). $\mu \in (0,1)$ is a virtual damping factor that ensures the robot does not continue moving indefinitely after force is applied in the future $Z-1$ time steps. This force compliance term in the formula ensures that over the next $Z-1$ time steps, the robot's velocity aligns as closely as possible with the direction of the force-compliant velocity $v_{\text{tgt}_N}$, while maintaining smooth transitions. When the user does not apply wrench exceeding a certain threshold, the force-compliant velocity $v_{\text{tgt}_N} = \mathbf{0}$, causing FC-MPC to degrade into a standard MPC. In this case, this term ensures that the robot's velocity remains as minimal as possible and transitions smoothly. $p(x_{t+Z|t}) := \left\| x_{t+N|t} - x^d_{t+N|t} \right\|_{\mathbf{Q}'}$ is the terminal cost, and $q(x_k) := \left\| x_k - x^d_k \right\|_{\mathbf{Q}}$ is the stage cost. These two terms ensure that the local planner follows the navigation path generated by the A* global planner. By adjusting $\mathbf{R}$, $\mathbf{S}$, and $\mathbf{Q}(\mathbf{Q}')$, the importance of force compliance (FC), velocity smoothness, and global navigation (GN) can be fine-tuned. When the corresponding weight matrix is set to a zero matrix, the associated functionality is disabled, allowing for various functional combinations.

Equation (13b) represents the system kinematic constraints. Equation (13c) ensures that the state variables and inputs remain within the allowable range of the actual system, where $\chi \subseteq \mathbb{R}^3$ and $U \subseteq \mathbb{R}^3$ denote the feasible sets of robot states and control inputs, respectively. Specifically,

$$\chi = \left\{ x_k \mid (x_k, y_k) \in \mathbb{R}^2, \theta_k \in [-\pi, \pi] \right\}, \tag{14}$$

$$U = \left\{ u_k \mid |v_{x_k}| \le v_{\max}, |v_{y_k}| \le v_{\max}, |\omega_{z_k}| \le \omega_{\max} \right\}, \tag{15}$$

where $v_{\max} > 0$ and $\omega_{\max} > 0$ denote the robot's maximum linear and angular velocities, respectively. (13d) defines the initial values, and specifies the terminal value range. Finally, (13e) is a **hard** constraint that ensures the safety of both the guide robot and the blind user. A **soft**-constrained version is proposed as an improvement to increase feasibility. The revised soft constraint and the definitions of variables $i$, $j$, $m$ and $\beta$ will be further explained in inequality constraints (20).

*B. Obstacle Perception and Obstacle Avoidance*

*(1) Control Barrier Functions for robot and user*

During the guide process, to ensure the safety of both the blind user and the robot, collision-free obstacle avoidance is required. The combination of MPC and Control Barrier Function (CBF) has been proven effective in ensuring that the robot can successfully avoid obstacles during navigation. Here, we consider the safety sets of both the robot and the blind user to build the guide safety set.

As described in Section III.B(3), the parameters of the surrounding obstacles can be represented by

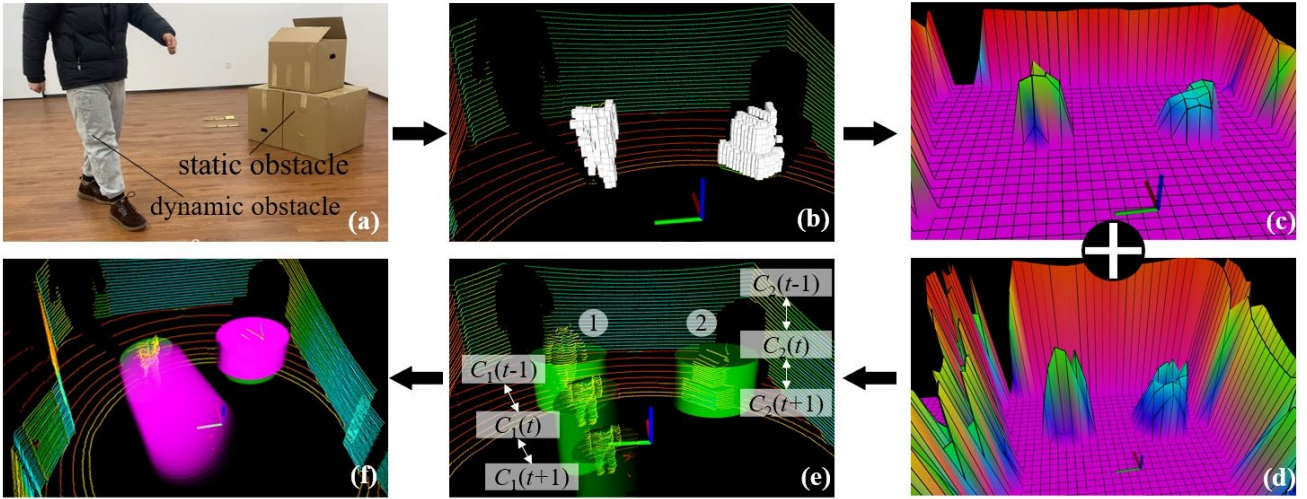

Fig. 5. An example for the entire obstacle perception process. (a) shows a pedestrian and a pile of cardboard boxes as dynamic and static obstacles respectively. (b) is the point cloud of the environment (small colorful points) and the obstacles (white cubic points). (c) and (d) are grid map visualization of the elevation matrix and the gradient matrix respectively. (e) is the result of the clustering and obstacle modeling. (f) Visualization of the predicted obstacle trajectory.

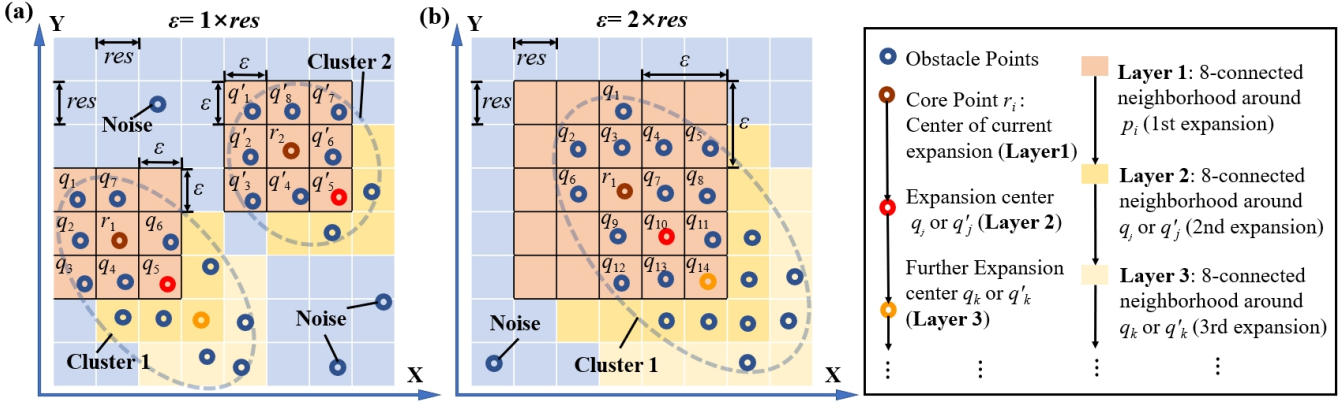

Fig. 6. Illustration of the $k$-order 8-connected grid-based DBSCAN clustering process, (a) is for $k$ =1 and (b) is for $k$ =2. Grid width equals $\varepsilon = k \cdot res$, which also determines the neighborhood radius. The figures demonstrate the recursive neighborhood expansion mechanism. The algorithm starts from a core point $r_i$ (Layer 1), expands to its $(2k+1) \times (2k+1)$ 8-connected neighborhood, and selects unvisited neighbors $q_i$ or $q'_i$ as new expansion centers (Layer 2), then continues further expansion (Layer 3) until all density-connected points are labeled.

$\boldsymbol{\psi} = \begin{bmatrix} x_{\text{obs}} & y_{\text{obs}} & a & b & \mathcal{G} \end{bmatrix}^T$. The robot's or the user's pose vector can be denoted as $\boldsymbol{x} = \begin{bmatrix} x & y & \theta \end{bmatrix}^T \in \mathbb{R}^3$. They are concatenated to form the safety state vector $\boldsymbol{X} = \begin{bmatrix} \boldsymbol{x} & \boldsymbol{\psi} \end{bmatrix}^T$ for either the robot or the user. Then the safety sets of the robot or the user $S_i$ are defined as:

$$S_i = \left\{ \boldsymbol{X} \in \Xi_i : h_i \left( \boldsymbol{X} \right) \geq 0 \right\} \quad (16)$$

, where $i$=1 is for the robot and $i$=2 is for the user. $\Xi_i$ represents the admissible range of values for the safety state vector of either the robot or the user. $h_i \left( \boldsymbol{X} \right)$ is the control barrier function.

When considering the safety of both the blind user and the guide robot, we can take the intersection of their respective safety sets to obtain the guide safety set:

$$S_{\text{guide}} = S_1 \bigcap S_2 . \quad (17)$$

.As illustrated in Fig. 4, we model obstacles as Minimum Bounding Ellipses (MBEs). Suppose at time $t$ there are $m$

MBEs, with the position of the $j$-th ellipse $O_j(t)$ given by $\boldsymbol{p}_{obs}^j(t) = \begin{bmatrix} x_{obs}^j(t) & y_{obs}^j(t) \end{bmatrix}^T$, where $j$=1, 2, ..., $m$. The position of the robot at time $t$ is $\boldsymbol{p}_{\text{robot}}(t) = \begin{bmatrix} x(t) & y(t) \end{bmatrix}^T$. The blind user's relative position is not fixed, but since the connecting rod is rigid and the distance $s$ from the grasping point to the robot's center varies very little in practice, we can approximate the user's average position as:

$$\boldsymbol{p}_{\text{human}}(t) = \boldsymbol{p}_{\text{robot}}(t) - s \cdot \begin{bmatrix} \cos\theta(t) & \sin\theta(t) \end{bmatrix}^T . \quad (18)$$

According to the prediction in Section III.B (3), obstacles can be modeled as multiple MBEs. Let the subscript $t+k|t$ be denoted as $k$. The position of the $j$-th ellipse at time $k$ is $\boldsymbol{p}_{\text{obs}}^j(k)$. The vector $\boldsymbol{p}_{\text{obs}}^j(k) - \boldsymbol{p}_{\text{robot}}(k)$ and the vector $\boldsymbol{p}_{\text{obs}}^j(k) - \boldsymbol{p}_{\text{human}}(k)$ form line segments that intersect the ellipse at points $P_k^j$ and $Q_k^j$, respectively. Based on geometric relationships, we can calculate the distances from the center of the MBEs to the intersection points $P_k^j$ and $Q_k^j$ at time $k$, which are denoted as $d_1^j(k)$ and $d_2^j(k)$. Thus, the Control Barrier Function for both the guide robot and the blind user are respectively defined as follows:

$$\begin{cases} h_1^j(\boldsymbol{X}_k) = \left\| \boldsymbol{p}_{\text{robot}}(k) - \boldsymbol{p}_{\text{obs}}^j(k) \right\|_2 - d_1^j(k) - d_{\text{safe}_1} \\ h_2^j(\boldsymbol{X}_k) = \left\| \boldsymbol{p}_{\text{human}}(k) - \boldsymbol{p}_{\text{obs}}^j(k) \right\|_2 - d_2^j(k) - d_{\text{safe}_2} \end{cases} \quad (19)$$

TABLE I
COMPARISON OF COMPUTATION TIMES FOR THREE TYPES OF DBSCAN METHODS

| $N_m$ | DBSCAN | | | | G-DBSCAN | | | | Ours | | | |
|---|---|---|---|---|---|---|---|---|---|---|---|---|
| | DBI | SIL | $\overline{t_1}$ (ms) | $\sigma_1$ (ms) | DBI | SIL | $\overline{t_2}$ (ms) | $\sigma_2$ (ms) | DBI | SIL | $\overline{t_3}$ (ms) | $\sigma_3$ (ms) |
| 150×150 | 0.46 | 0.795 | 19 | 0.1 | 0.46 | 0.789 | 12 | 0.06 | 0.47 | 0.741 | 8 | 0.04 |
| 300×300 | 0.41 | 0.782 | 95 | 2.3 | 0.53 | 0.715 | 30 | 2.4 | 0.54 | 0.732 | 25 | 2.6 |
| 450×450 | 0.42 | 0.701 | 280 | 37.0 | 0.49 | 0.682 | 78 | 6.0 | 0.48 | 0.688 | 61 | 4.2 |
| 600×600 | 0.46 | 0.763 | 670 | 74.1 | 0.49 | 0.701 | 124 | 14.2 | 0.52 | 0.681 | 95 | 8.0 |
| 750×750 | 0.48 | 0.778 | 1838 | 180.5 | 0.52 | 0.742 | 305 | 28.5 | 0.51 | 0.732 | 168 | 14.2 |
| 900×900 | 0.50 | 0.792 | 4055 | 250.3 | 0.50 | 0.751 | 585 | 49.0 | 0.55 | 0.725 | 343 | 24.1 |

, where $d_{\text{safe}_1}$ and $d_{\text{safe}_2}$ are the safety distances set for the guide robot and the blind user, respectively, based on safety requirements. For discrete systems, to ensure safety, we got (13e) as **hard** inequality constraints. However, this will bring feasibility problems when facing complex environment and multiple high speed dynamic obstacles. Temporary and penalized violations of the safety constraints is crucial in scenarios such as:

• **Obstacle intrusion**: When some dynamic obstacles enter the safety region unexpectedly ( $h_i^j(\boldsymbol{X}_k) < 0$ ), including cases where the robot, the user, or both violate their safety sets, a feasible control input may still be found, which will steer the system back to the safe set.

• **Conflicting safety sets**: If the robot and user's safe regions become disjoint (i.e., $S_1 \bigcap S_2 = \varnothing$ ), the optimizer can minimally violate some constraints based on task priority.

To achieve feasibility in the scenarios above, we relax the original hard robot-user control barrier function constraints by introducing non-negative slack variables $\delta_{i,k} \geq 0$. The following inequality constraints (20) are the safety **soft** constraints of the $j$-th ellipse at time step $k$:

$$h_i^j(\boldsymbol{X}_{k+1}) \geq (1-\beta) h_i^j(\boldsymbol{X}_k) - \delta_{i,k}, \begin{cases} \beta \in (0,1] \\ k = 0,1,...,Z-1 \\ i = 1,2 \\ j = 1, 2, ..., m \end{cases} \quad (20)$$

, where $i$=1 denotes the robot and $i$=2 denotes the user. $\beta$ represents a scaling factor that controls the extent of obstacle avoidance.

We define the full slack vector over the prediction horizon as:

$$\boldsymbol{\delta} := \left\{ \delta_{i,k} \right\}_{i=1,2; k=0,1,...,Z} \in \mathbb{R}^{2Z}. \quad (21)$$

This vector is treated as part of the optimization variable set and is heavily penalized in the cost function to minimize safety constraint violations when hard feasibility cannot be maintained. The updated MPC objective is:

$$\min_{\boldsymbol{u},\boldsymbol{\delta}} p(\boldsymbol{x}_Z) + \sum_{k=0}^{Z-1} \left( q(\boldsymbol{x}_k) + v(\boldsymbol{u}_k) \right) + \sum_{k=0}^{Z-1} \sum_{i=1}^{2} \kappa_i \delta_{i,k}^2 \quad (22)$$

, where the last term penalizes violations of the safety constraints. The weight $\kappa_i > 0$ is safety coefficient for the robot and the user, from which we can adjust the tolerance for risk. In practice, $\kappa_i$ is set much larger than the elements of the weight matrices $\boldsymbol{Q}$, $\boldsymbol{Q'}$, $\boldsymbol{R}$, and $\boldsymbol{S}$ to ensure that the safety constraints are only violated as a last resort. We can adjust the values of $\kappa_1 / \kappa_2$ to prioritize either the robot's or the user's safety. Typically, we set $\kappa_2 / \kappa_1 \gg 1$ to ensure that user safety is prioritized when conflicts arise between the robot's and user's safety sets.

The formulation in Eq. (20) ensures discrete-time forward invariance in a relaxed but formally verifiable sense. Specifically, for each robot or user safety constraint, if the slack variable satisfies $\delta_{i,k} = 0$ and $h_i^j(\boldsymbol{X}_k) > 0$, then the inequality becomes:

$$h_i^j(\boldsymbol{X}_{k+1}) \geq (1-\beta) h_i^j(\boldsymbol{X}_k) > 0, \quad (23)$$

which guarantees that the safety constraint remains satisfied at the next step—i.e., forward invariance holds in the standard discrete-time CBF sense. When $h_i^j(\boldsymbol{X}_k) < 0$, or when the robot's and user's safety sets become disjoint (i.e., $S_1 \bigcap S_2 = \varnothing$ ), the asymmetric slack penalties allows the optimizer to maintain feasibility and prioritize user safety. The penalization also encourages finite-time recovery, ensuring that the system is softly guided back into the safe set within a few steps.

**Algorithm 1** EightConnectedDBSCAN(eps, minPts, D)

---

mark all points in D as unvisited
cluster_id ← 1
initialize 8-connected grid index for D with resolution eps

**for each** unvisited point $r$ in D **do**
  N ← find_neighbors($r$, eps)
  **if** |N| < minPts **then**
    mark $r$ as noise
  **else**
    assign cluster_id to $r$ and all points in N
    queue ← all unvisited points in N
    **while** queue is not empty **do**
      $q$ ← remove first point from queue
      Nq ← find_neighbors($q$, eps)
      **if** |Nq| ≥ minPts **then**
        **for each** point $p$ in Nq **do**
          **if** $p$ is unvisited **then**
            assign cluster_id to $p$
            add $p$ to queue
      **end if**
      mark $q$ as visited
    **end while**
    mark $r$ as visited
    cluster_id ← cluster_id + 1
  **end if**
**end for**

Output all points in D with their cluster labels or noise

**function** find_neighbors($r$, eps):
  neighbors ← []
  **for each** point $q$ in 8-connected grids around $r$ **do**
    add $q$ to neighbors
  **return** neighbors

---

The presented Robot-User CBFs enable us to simultaneously ensure the safety of both the user and the robot while maintaining efficient navigation.

*(2) Binary Map Construction and Obstacle Clustering*

The obstacle perception process is shown in Fig. 5. The guide robot scans the surrounding point cloud map using LiDAR. The point cloud is then processed through filtering and down-sampling to generate a 2D binary matrix map. Subsequently, a clustering algorithm is applied to group obstacle points within a certain area and assign them unique ID labels. Previous work used the traditional DBSCAN method for clustering [2]; however, its computational complexity is high for resource-constrained robots. Inspired by work [21], we adopt an Eight-Way Connected DBSCAN Clustering Method, which significantly outperforms traditional DBSCAN method in terms of computational efficiency. After clustering, we model the surrounding obstacles as Minimum Bounding Ellipses (MBEs) and use Kalman filtering to estimate and predict the obstacles and their future trajectories.

We use the scenario illustrated in Fig. 5(a) as an example to visualize the complete obstacle perception process. A pedestrian walking at approximately 1.5 m/s and a pile of cardboard boxes were selected as dynamic and static obstacles, respectively.

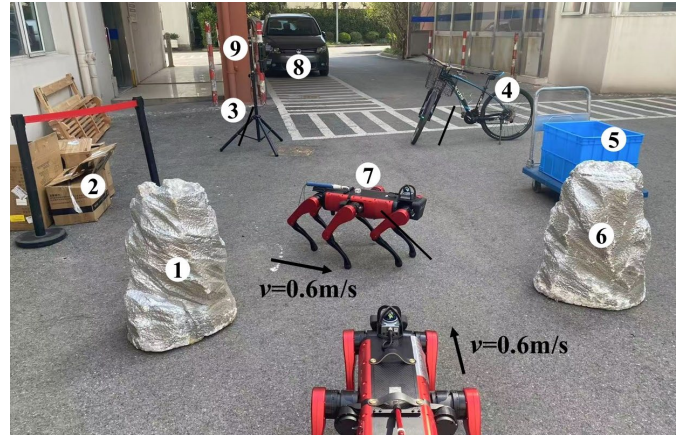

Fig. 7. In the clustering experiment scenario, the guide robot repeatedly crosses a variety of static and dynamic obstacles at a speed of 0.6 m/s, and the dynamic obstacle moves at the same speed. Clusters are marked with numbers.

We selected the point cloud within a certain range of interest around the robot, which is called the local map. The 3D point cloud map is then voxelized into a 2D matrix $\mathbf{Q}_{i \times j}$. The value of the element $q_{i \times j}$ is defined as the maximum z-coordinate value within the neighborhood corresponding to $(i, j)$. The white cubic point cloud in Fig. 5(b) represents all the highest points within their respective neighborhoods in the local map. Then, the gradient matrices for the grid map matrix $\mathbf{Q}_{i \times j}$ are constructed using the Sobel operator to obtain the gradient matrix $\mathbf{S}_{i \times j}$. The grid map matrix is then binarized into matrix $\mathbf{G}_{i \times j}$, with the binarization method as follows:

$$g_{ij} = \left( q_{ij} > q_{\text{th}} \right) \oplus \left( s_{ij} > s_{\text{th}} \right), \tag{24}$$

where "$\oplus$" is the logical OR operator, and $g_{ij}$, $q_{ij}$ and $s_{ij}$ are arbitrary elements from the matrices $\mathbf{G}_{i \times j}$, $\mathbf{Q}_{i \times j}$ and $\mathbf{S}_{i \times j}$, respectively. The thresholds $q_{\text{th}}$ and $s_{\text{th}}$ represent the height threshold for determining an obstacle and the first-order gradient threshold, respectively. Fig. 5(c) and Fig. 5(d) respectively visualize the elevation matrix $\mathbf{Q}_{i \times j}$ and the gradient matrix $\mathbf{S}_{i \times j}$, demonstrating that the above method can effectively identify obstacles.

After obtaining the binary matrix $\mathbf{G}_{i \times j}$, an Eight-Way Connected DBSCAN Clustering Method is applied to cluster obstacle regions.

In traditional DBSCAN, for each point $r_i \in \mathcal{D}$, the neighborhood is defined in continuous Euclidean space as:

$$\mathcal{N}_{\varepsilon} \left( r_i \right) = \left\{ r_j \in \mathcal{D} \mid \left\| r_j - r_i \right\|_2 \leq \varepsilon \right\}, \tag{25}$$

and $r_i$ is considered a core point if $\left| \mathcal{N}_{\varepsilon} \left( r_i \right) \right| \geq$ minPts. The minPts is the minimum number of points of a cluster. Clusters are formed by recursively expanding from each core point through all density-connected neighbors. The algorithm typically requires pairwise distance computations, leading to O($N^2$) time complexity in the worst case.

In contrast, our Grid-DBSCAN algorithm operates in a discretized occupancy grid. Each obstacle point $r_i$ has already been associated with a unique grid cell, derived from a predefined resolution $res$. The clustering radius is defined as an integer multiple of the resolution, i.e., $\varepsilon = k \cdot res, k \in \mathbb{Z}^+$. The parameter $k$ controls the spatial scale of density reachability. Adjusting $k$ plays the same role as tuning $\varepsilon$ in traditional DBSCAN. Accordingly, we define the neighborhood of a point $r_i$ by examining the grid cells in the $(2k+1) \times (2k+1)$ 8-connected region centered at $r_i$. Formally:

$$\mathcal{N}_\varepsilon(r_i) = \{r_j \in \mathcal{D} \mid r_j \in \text{Grid}_k(r_i)\}, \tag{26}$$

where $\text{Grid}_k(r_i)$ denotes the set of neighboring grid indices within $k$-order 8-connectivity. A point is considered a core point if $|\mathcal{N}_\varepsilon(r_i)| \geq \text{minPts}$, and cluster expansion is recursively performed through adjacent grid cells.

This grid-based formulation avoids distance computations entirely and enables constant-time neighborhood lookup. As a result, it preserves the density-based semantics of DBSCAN while significantly reducing the computational cost from O($N^2$) to approximately O($N$). This makes it particularly suitable for large-scale point cloud maps on resource-constrained robotic systems like guide robot.

The recursive manner of clustering proceeds is as follows. Starting from a core point $p_i$, all reachable neighbors within 8-connected cells are labeled with the same cluster ID. If a neighbor $q_i$ itself has sufficient(>=minPts) neighbors, it becomes a new expansion center, leading to multi-layer propagation of density-connected regions. This process is visualized in Fig. 6, and the detailed steps are presented in **Algorithm 1**, in which we take $\varepsilon = res$ as an example.

To evaluate the performance of the Eight-Way Connected DBSCAN, we compare it with the traditional DBSCAN adopted by [2] and a fast G-DBSCAN method [23] in terms of both computation time and clustering accuracy. We applied the above clustering method to the scene shown in Fig. 7, which contains multiple static and dynamic obstacles. The local map was configured with a fixed size of 15 meters by 15 meters, with the number of sample grid points $N_m$ determined by varying resolutions. The number of grid points along each edge of the local map was incremented at equal intervals.

For the computation time comparison, we run all clustering methods on a Jetson Xavier NX platform, recording the mean computation times $\overline{t_1}$, $\overline{t_2}$, and $\overline{t_3}$ along with their corresponding standard deviations $\sigma_1$, $\sigma_2$, and $\sigma_3$. For the clustering accuracy, we further evaluate clustering compactness and separation by two metrics: Davies-Bouldin Index (DBI) and Silhouette Score (SIL). DBI measures the average "worst-case similarity" between each cluster and its most similar neighbor cluster. It is defined as:

$$\text{DBI} = \frac{1}{n} \sum_{i=1}^{n} \max_{j \neq i} \left( \frac{S_i + S_j}{M_{ij}} \right), \tag{27}$$

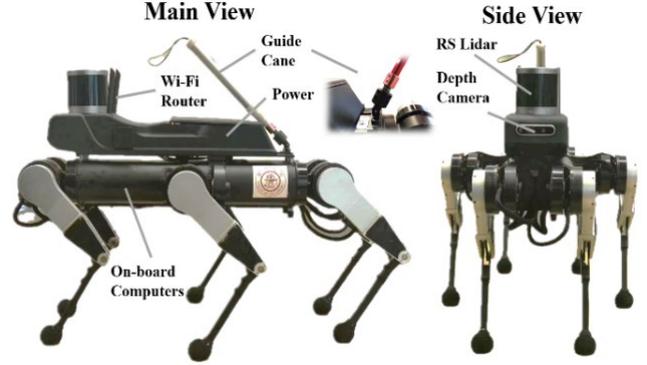

**Fig. 8.** HexGuide Hexapod Guide Robot Hardware Platform

, where $n$ is the number of clusters, $S_i$ is the average distance between each point in cluster $i$ and the centroid of cluster $i$ (intra-cluster dispersion), and $M_{ij}$ is the Euclidean distance between the centroids of clusters $i$ and $j$. The DBI is non-negative, where lower values suggest more compact and well-separated clusters. The SIL is defined as:

$$\text{SIL} = \frac{1}{N_m} \sum_{i=1}^{N_m} \frac{\lambda_i - \rho_i}{\max(\lambda_i, \rho_i)}, \tag{28}$$

, where $\rho_i$ is the mean intra-cluster distance of sample point $i$, and $\lambda_i$ is the mean nearest-cluster distance of point $i$. The SIL ranges from -1 to 1, a value close to 1 indicates well-separated and dense clusters.

Table I summarizes the performance comparison. Despite complex obstacle scenarios, our Eight-Way Connected DBSCAN achieves a substantial speedup over traditional DBSCAN and a moderate improvement over G-DBSCAN. As grid resolution increases, the efficiency advantage becomes more apparent—reaching up to 11.8× speedup. This is achieved by sacrificing a small degree of clustering accuracy, as the 8-connected grid approach may miss points near cluster boundaries that lie across adjacent cells but are within the Euclidean radius, leading to minor over-segmentation or noise. However, the increase in DBI and the decrease in SIL remain within acceptable bounds, indicating no significant loss in clustering quality. Furthermore, such edge misclassifications can be mitigated in practice by enlarging the CBF safety margin. In contrast, traditional DBSCAN takes up to 4.055

## TABLE II
### Experimental Scenarios and Corresponding Features

| No. | Guide Scenarios | FC | GN | OA |
|-----|-----------------|-----|-----|-----|
| 1 | Pure Force Interaction Control Locomotion | √ | × | × |
| 2 | Force-adaptive autonomous navigation | √ | √ | × |
| 3 | Force Interaction Control Locomotion with Obstacle Avoidance | √ | × | √ |
| 4 | Dynamic obstacle avoidance in force-compliant autonomous navigation | √ | √ | √ |
| 5 | Long Distance Indoor and Outdoor Guide | √ | √ | √ |

Notes: 'FC' stands for Force Compliance, 'GN' stands for Global Navigation while 'OA' stands for Obstacle Avoidance.'

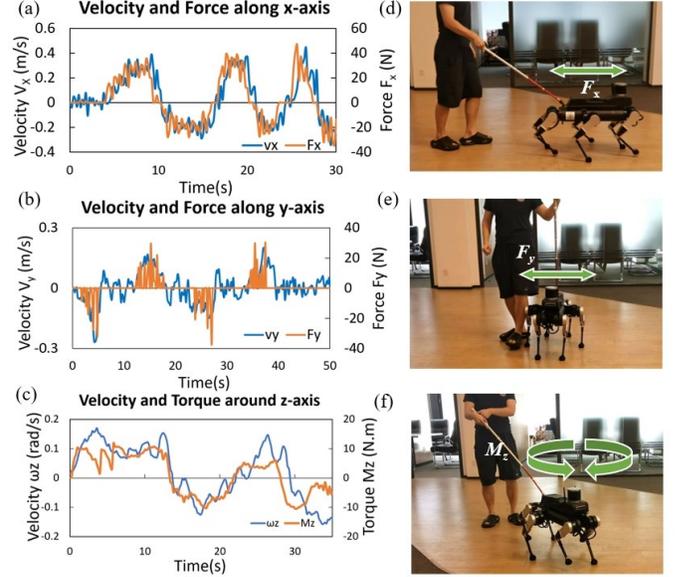

Fig. 9. Force Compliance under Pure Force Interaction Control. (a), (b), and (c) show the velocity commands of the three directions and the corresponding forces or moments. (d), (e), and (f) are schematic photos of the experiment.

seconds on embedded hardware, and when factoring in downstream processes like prediction and FC-MPC, the overall local planning frequency may drop to 0.25~4 Hz, which is insufficient for real-time navigation. Thus, our method offers a practical trade-off between accuracy and efficiency.

As the consequence of clustering algorithm, we got $n$ obstacle clusters $C = \{C_1, C_2, …, C_j, …, C_n\}$. Considering the discrete time stamp $t$ of different frames, the cluster will be denoted as $C_1(t)$, $C_2(t)$, …, and $C_n(t)$.

### (3) Obstacle Modeling and Prediction

After clustering, each cluster $C_j$ we got is then parameterized as the Minimum Bounding Ellipses (MBEs). The obstacle $C_j$ is parameterized as an ellipse $\varepsilon_i$, described by its center coordinates $\boldsymbol{p}_{obs}^j = \left[ x_{obs}^j \quad y_{obs}^j \right]^T$, semi-major axis $a_j$, semi-minor axis $b_j$, and rotation angle $\theta_j$ as follows:

$$\varepsilon_j = \left( \boldsymbol{p}_{obs}^j \quad a_j \quad b_j \quad \theta_j \right)^T, j = 1,2,...,n \quad (29)$$

The Kuhn-Munkres algorithm [43] is applied to achieve obstacle matching between frames. To associate and track obstacles across consecutive frames, the Euclidean distance between the center of each ellipse in the previous frame $t$-1 and the current frame $t$ is calculated as:

$$d_{mn} = \left\| \boldsymbol{p}_{obs}^j(t) - \boldsymbol{p}_{obs}^j(t-1) \right\|_2 . \quad (30)$$

Then, the association matrix $\mathbf{B}$ is defined as:

$$b_{mn} = \begin{cases} 1, & \text{if } d_{mn} \leq d_{\max} \\ 0, & \text{otherwise} \end{cases} . \quad (31)$$

Through the association matrix B, we can achieve inter-frame matching of the $j$-th obstacle. Newly appearing obstacles are assigned new labels. The results are shown in Fig. 5(e).

After performing inter-frame obstacle matching and labeling, Kalman filtering (KF) is applied to the parameterized MBEs for state estimation, in order to achieve trajectory prediction of dynamic obstacles.

The state variable of obstacles is defined as $\boldsymbol{\zeta} = \left[ x_{obs} \quad y_{obs} \quad \dot{x}_{obs} \quad \dot{y}_{obs} \quad \ddot{x}_{obs} \quad \ddot{y}_{obs} \quad a \quad b \quad \theta \right]^T$.

$\boldsymbol{z} = \left[ x_{obs} \quad y_{obs} \quad a \quad b \quad \theta \right]^T$ is the observable variable. Then the state transition matrix $\mathbf{A}$ is defined as follows:

$$\mathbf{A} = \begin{bmatrix} 1 & 0 & T & 0 & \dfrac{T^2}{2} & 0 & \\ 0 & 1 & 0 & T & 0 & \dfrac{T^2}{2} & \mathbf{0}_{4\times3} \\ 0 & 0 & 1 & 0 & T & 0 & \\ 0 & 0 & 0 & 1 & 0 & T & \\ & & \mathbf{0}_{5\times6} & & & & \mathbf{I}_{5\times3} \end{bmatrix} . \quad (32)$$

Based on the system conditions, the system noise matrix and the measurement noise matrix are defined, thereby obtaining the state trajectory $\boldsymbol{\zeta}_k$ of the dynamic obstacle over the next $N_{obs}$ time steps, where $k = 1,2,...,N_{obs}$. The predicted trajectories of the obstacles are visualized in purple ellipses in Fig. 5(f). These trajectories are used to construct the Control Barrier Functions for each time step.

## IV. EXPERIMENTS

### A. Hardware Platform

As shown in Fig. 8, our algorithm is deployed on the HexGuide hexapod guide robot, equipped with 6 legs, 18 motors, and a guide cane connected through a rotating joint. The Nvidia Jetson Xavier NX onboard computer handles ROS-based localization, force estimation, interaction, and FC-MPC. Motor commands are transmitted via Wi-Fi to control the 18 joint motors. The robot is equipped with a 32-line

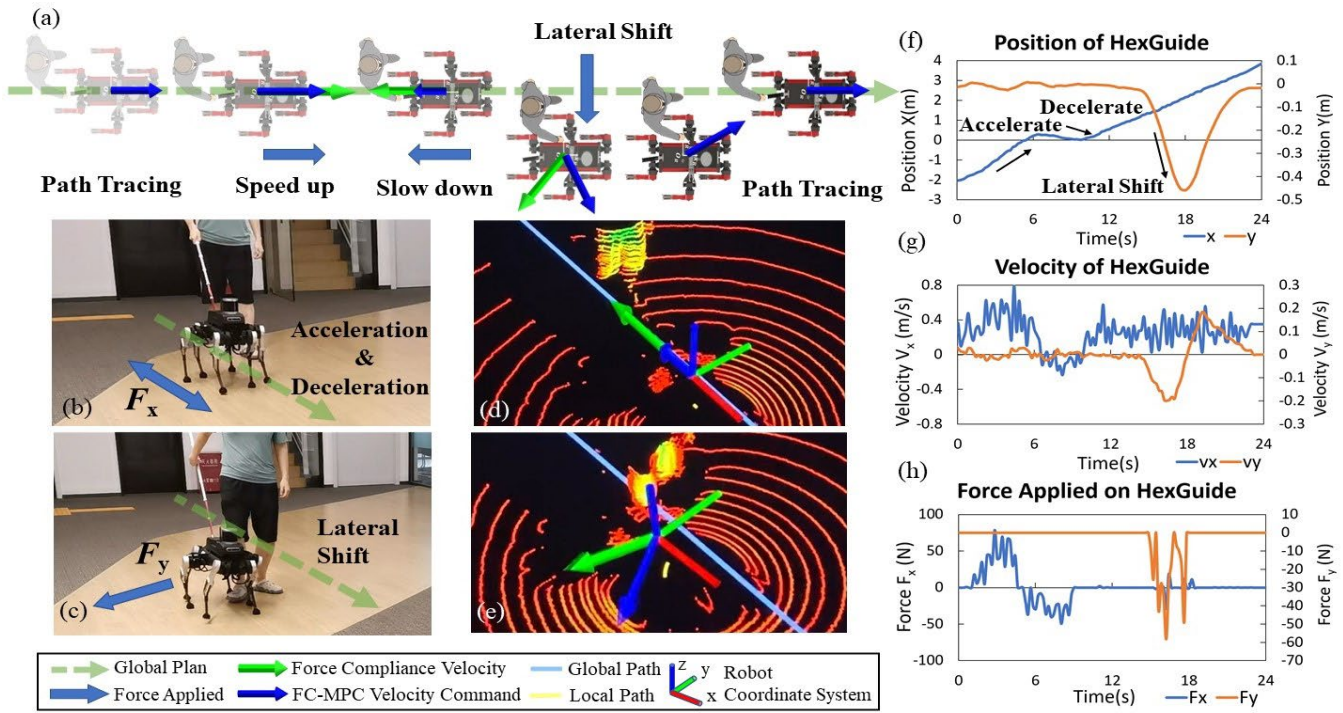

Fig. 10. Force-Compliant Autonomous Navigation Experiment. (a) The user controls the robot through force for acceleration, deceleration, and lateral shifts, with the robot returning to the global path when no force is applied. (b) The user pushes with a cane to accelerate and pulls to decelerate. (c) Lateral movement is achieved by applying force in the y direction. (d) and (e) are the visualization of the force compliance velocity and the velocity command computed by FC-MPC, during deceleration and lateral shift process respectively. (f), (g) and (h) show the robot's position and velocity and force along the x and y axes.

RoboSense LiDAR, collecting point cloud data at 10Hz. Local perception runs at 10Hz, and the obstacle avoidance algorithm operates at 8-10Hz.

### B. Experimental Scenarios

In order to fully simulate the use of guide dogs in real life, we have identified five key questions, each corresponding to a distinct scenario for guide robot:

**Q1:** As the foundation of two-way force interaction, can the guide robot effectively perceive the forces and torques applied by the user and respond with appropriate compliance?

**Q2:** During the robot's autonomous navigation process, can it still comply with external forces applied by the user, thus achieving two-way force interaction throughout the navigation process seamlessly?

**Q3:** In scenarios where the blind user directs the global movement by applying external forces, rather than relying on the robot's autonomous navigation, can the robot's local obstacle avoidance still function effectively to ensure the safety of both the user and the robot?

**Q4:** In scenarios where the global movement direction is mainly determined by the robot's autonomous navigation, can the local obstacle avoidance and force compliance still function effectively, ensuring the safety of both the user and the robot, even when facing dynamic obstacles?

**Q5:** Is the proposed method robust enough for real-world indoor and outdoor blind guiding applications?

As shown in Table II, through fine-tuning of FC-MPC and CBFs, we can freely combine the three functionalities—**F**orce **C**ompliance (**FC**), **G**lobal **N**avigation (**GN**), and **O**bstacle **A**voidance (**OA**)—to meet the requirements of different guide scenarios, thereby addressing the above four questions. Experimental results demonstrate the generality and effectiveness of our proposed method.

#### (1) Pure Force Interaction Control Locomotion

To address **Q1**, we conducted an ablation study, retaining only the force-compliance component to intuitively showcase the six-legged guide robot's ability to estimate and comply with external forces and moments. At this stage, the coefficient matrices $\mathbf{Q}$ and $\mathbf{Q}'$ of the global navigation (GN) component are set to zero, and the CBF constraints are not applied.

Under this pure force-interaction scenario, the robot fully complies with user-applied forces in the $x$, $y$ and moments around the $z$-axes, relying solely on force compliance for navigation. In Fig. 9, we plot the velocity command $v_x$, $v_y$ and $\omega_z$ computed with FC-MPC, which will be send to the upper controller. We also plot the external force $F_x$, $F_y$ and moments $M_z$ estimated with the method mentioned in Section III.A. Those figures illustrate how the robot responds to forces: it moves when the force accumulates and gradually comes to a stop once the force is removed.

#### (2) Force-Compliant Autonomous Navigation

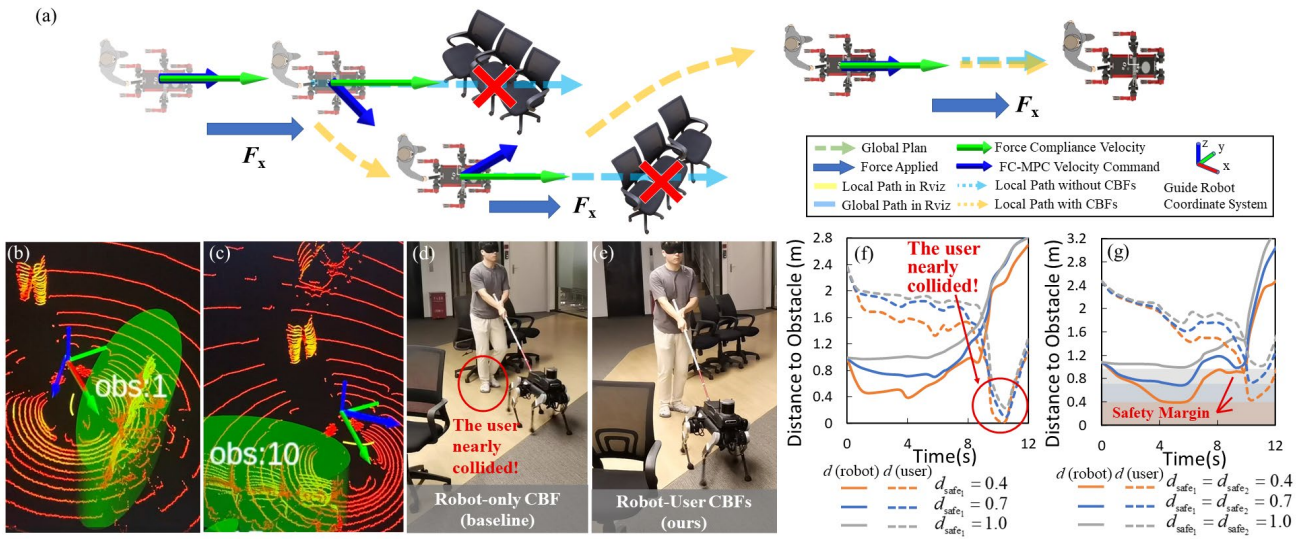

Fig. 11. Obstacle avoidance experiment under force interaction control locomotion. (a) The user pushes the robot forward; with CBF activated, obstacle avoidance is enabled. (b) and (c) are Rviz visualization screenshots of force compliance velocity, velocity command, and local plan when avoiding the first and second row of obstacles. (d) and (e) compare robot-only CBF and Robot-User CBFs. The latter accounts for both user and robot. (f) and (g) show the minimum distance between the robot, user, and obstacles under different safe distances. With robot-only CBF, the user nearly collides, while with Robot-User CBFs, both stay within a safe range.

To address **Q2**, we designed a scenario where the robot initially follows a global path autonomously. During navigation, pushing or pulling forces $F_x$ regulate its forward speed to adjust walking rhythm. Later, a lateral force $F_y$ is applied by physically contacting the robot, altering its direction and achieving lateral displacement. This mimics how users physically interact with guide dogs when deviating from a planned route. This function is realized by disabling CBF constraints and fine-tuning the FC-MPC coefficient matrices. Fig. 10(a), (b), and (c) illustrate this scenario comprehensively.

Fig. 10(d) and (e) show Rviz visualizations during deceleration and lateral displacement. The green and blue arrows respectively represent the force compliance velocity $v_{\text{tgt}_N}$ and the two-dimensional vector $\begin{bmatrix} v_x & v_y \end{bmatrix}^T$ derived from the velocity command calculated by FC-MPC, which is sent to the upper controller. When the user does not apply any external force, the velocity command points almost directly forward. When pushing forces are applied along the forward direction, $v_{\text{tgt}_N}$ also aligns with the forward direction, and the velocity command increases in magnitude along this direction. Conversely, when pulling forces are applied opposite to the forward direction, $v_{\text{tgt}_N}$ reverses, and the velocity command decreases to zero or even reverses. When lateral forces are applied, the velocity command direction lies between $v_{\text{tgt}_N}$ and the forward direction.

Summarizing these patterns, the velocity command vector $\begin{bmatrix} v_x & v_y \end{bmatrix}^T$ computed by FC-MPC can be approximately interpreted as the vector sum of $v_{\text{tgt}_N}$ and the trajectory tracking velocity without external forces. Adjusting the matrices $\mathbf{R}$ and $\mathbf{Q}(\mathbf{Q'})$ is analogous to scaling these two vectors with respective coefficients. Although this is just a conceptual interpretation rather than a strict mathematical equivalence, it effectively reveals the principles of FC-MPC.

Figures 10(f), (g), and (h) detail the robot's position, velocity, and the external forces experienced throughout the entire process. FC-MPC seamlessly integrates force compliance and autonomous navigation by setting the similarity between output velocities and $v_{\text{tgt}_N}$ as an MPC optimization objective, achieving a *smooth* fusion of force compliance and autonomous navigation.

(3) *Force Interaction Control Locomotion with Obstacle Avoidance*

**Q3** addresses a common scenario where the user determines the global path, while the guide robot handles local obstacle avoidance. To address **Q3**, we designed the experiment illustrated in Fig. 11(a). As shown in Fig. 11(a), the user continuously applies a forward pushing force to express the intention to move straight, but obstacles ahead prevent full compliance. The robot must therefore balance safety and compliance, avoiding collisions while staying as close as possible to the user's intended direction.

Our proposed method effectively satisfies this requirement by simply adding Robot-User CBFs constraints to the parameter settings of FC-MPC in the Pure Force Interaction Control Locomotion experiment mentioned in Section IV.B(1). This is achieved by setting $\kappa_i$ to a positive value much greater than the elements of the weighting matrices $\mathbf{Q}$, $\mathbf{Q'}$, $\mathbf{R}$, and $\mathbf{S}$. Fig. 11(b) and (c) show the force compliance velocity $v_{\text{tgt}_N}$, the velocity command $\begin{bmatrix} v_x & v_y \end{bmatrix}^T$ and the local path. While the force compliance velocity $v_{\text{tgt}_N}$ aligns with the

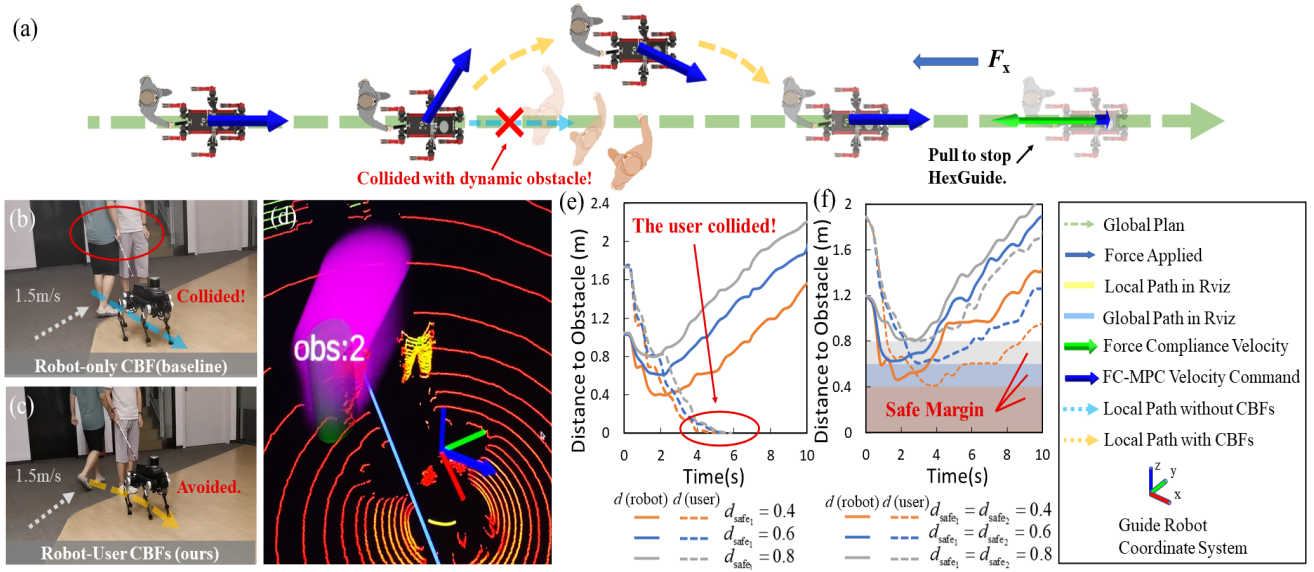

Fig. 12. Force-Compliant Autonomous Navigation with Dynamic Obstacle Avoidance Experiment. (a) is a diagram of the experiment designed for Q4. The robot guides the blind user to avoid the dynamic obstacle moving at 1.5 m/s. If the Robot-User CBFs are not used, there will be a collision. After avoiding the obstacle, the user will pull the robot to slow down. In (b) and (c), we compare the obstacle avoidance performance of robot-only CBF and Robot-User CBFs. (d) is the Rviz visualization of the obstacle avoidance process. (e) and (f) show minimum distances from the robot or user to the obstacle under different safe distance. The user collided with the pedestrian with robot-only CBF, while with Robot-User CBFs, both the robot and user remain safe.

<table>
<tr><th colspan="2">TABLE III<br>Success Rates of Different CBFs for Q3</th></tr>
</table>

| CBFs | Success Rates |
|---|---|
| robot-only (hard constraint) | 41.7% (10/24) |
| robot-user (hard constraint) | 62.5% (15/24) |
| robot-only (soft constraint) | 75.0% (18/24) |
| robot-user (soft constraint, ours) | **100.0%** (24/24) |

<table>
<tr><th colspan="2">TABLE IV<br>Success Rates of Different CBFs for Q4</th></tr>
</table>

| CBFs | Success Rates |
|---|---|
| robot-only (hard constraint) | 33.3% (10/30) |
| robot-user (hard constraint) | 50.0% (15/30) |
| robot-only (soft constraint) | 40.0% (12/30) |
| robot-user (soft constraint, ours) | **96.7%** (29/30) |

user's pushing direction, the MPC velocity command deviates slightly under CBF constraints to avoid collisions while staying as close to the intended direction as possible.

We conducted 25 trials and present the most representative cases in Fig. 11(d) and (e), comparing robot-only CBF ($\kappa_1 \neq 0, \kappa_2 = 0$) and our proposed Robot-User CBFs ($\kappa_1 \neq 0, \kappa_2 \neq 0$). The Robot-User CBFs maintain safe distances for both robot and user, whereas robot-only CBF may result in near-collisions for the user. Fig. 11(f) and (g) plot minimum distances between the robot or user and the nearest obstacle. We further evaluated the average performance of the Robot-User CBFs under different safety distance $d_{\text{safe}_i}$ settings. The figures illustrate how adjusting the safety distance changes the conservativeness of obstacle avoidance, allowing flexible system configuration.

Table III shows the success rates of all CBF strategies across 24 trials each. A trial fails if a collision occurs or if the system is infeasible for over 5 seconds. Results show that our soft-constrained Robot-User CBF achieves 100% success, effectively ensuring robot-user safety and avoiding local infeasibility.

(4) *Dynamic Obstacle Avoidance in Force-Compliant Autonomous Navigation*

**Q4** examines a real-world case where the robot must autonomously navigate while avoiding dynamic obstacles and maintaining force interaction with the user. This requires fine-tuning all FC-MPC matrices and adding CBFs to jointly ensure safety.

As shown in Fig. 12(a), we tested this scenario with a 1.5 m/s pedestrian approaching the user. After avoiding the obstacle, the user pulled the robot to simulate a cautious slowdown. Fig. 12(b) and (c) show that robot-only CBF fails to prevent collision, while Robot-User CBFs succeed. In Fig. 12(d), we provide an Rviz visualization during dynamic obstacle avoidance, where the velocity command, originally directed along the forward path, gains a lateral component under the constraints of the CBFs to prevent collisions.

In Fig. 12(e) and (f), we further plot minimum distances under different safety settings, confirming that Robot-User CBFs maintain safe margins for both robot and user.

Table IV reports results from 30 trials, with failure defined as collision or prolonged infeasibility. Our method consistently achieves better performance, as it explicitly treats user safety as a core optimization target rather than a secondary outcome of robot-only obstacle avoidance.

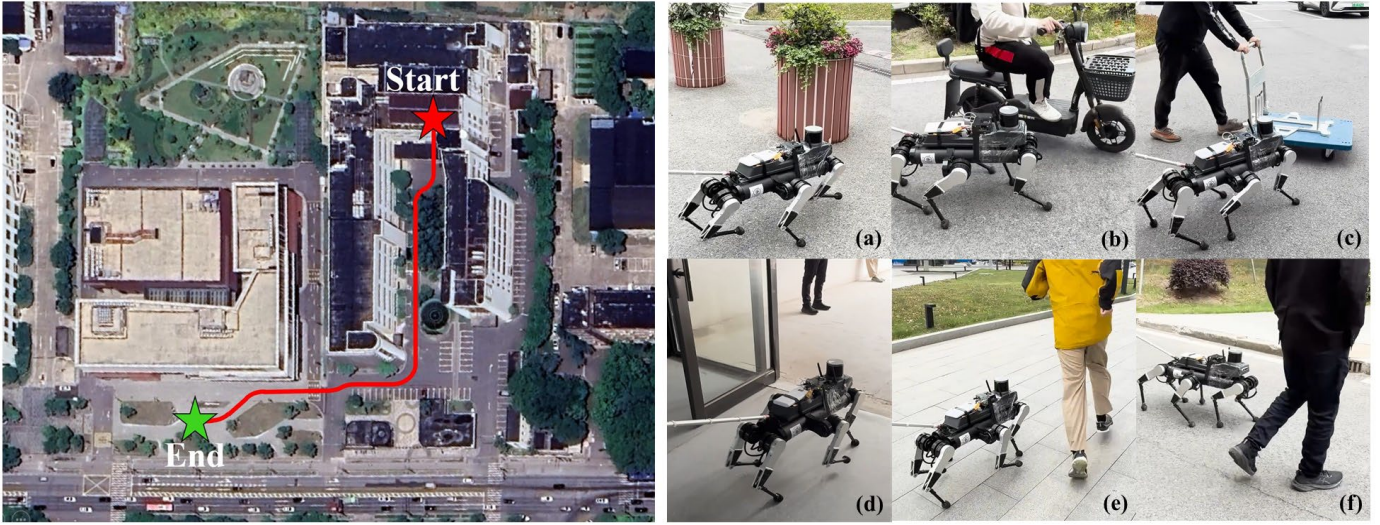

Fig. 13.    (a) shows the satellite map and global route of the long-distance guiding experiment, which lasted 5 minutes and 48 seconds and covered approximately 220 meters. Along the way, the robot encountered various static obstacles, including (a) a flower planter and (d) a glass door. Dynamic obstacles included (b) electric bikes moving at 2 m/s, (c) a person pushing a cart at 1.5 m/s, (e) a running pedestrian at 3 m/s, and (f) a person strolling at 0.8 m/s.

### (5) Long Distance Indoor and Outdoor Guide

To answer **Q5**, we conducted an uninterrupted long-distance guiding experiment lasting 5 minutes and 48 seconds, covering approximately 220 meters from an indoor to an outdoor environment, as shown in Fig. 13(a). As illustrated in Fig. 13(b), the route included both static obstacles and various dynamic obstacles moving at speeds ranging from 0.5 m/s to 3 m/s. This setup was designed to evaluate the robustness and reliability of the proposed FC-MPC and Robot-User CBFs in practical scenarios. The experiment video shows that the robot was able to handle diverse static and dynamic disturbances smoothly and successfully completed the guiding task.

### V. Conclusion

In this paper, we presented FC-MPC and adopted Robot-User CBFs as an effective local planner for safe, two-way interactive navigation for guide robot. We constructed a guide robot control framework centered on the proposed local planner, encompassing global localization and path planning, local obstacle perception, external force estimation and compliance, and a low-level walking controller. To enable deployment on computationally constrained robots without remote computing assistance, we utilized the Eight-Way Connected DBSCAN clustering method, reducing time complexity from $O(n^2)$ to approximately $O(n)$. This complete algorithmic framework has been successfully implemented on the HexGuide hexapod robot.

Experimental validation confirms the method's effectiveness in seamlessly integrating global autonomous navigation, force compliance and obstacle avoidance, offering a practical solution for enhancing the interactivity and safety of guide robots in dynamic, obstacle-rich environments.

The proposed framework is not limited to guide robots requiring force interaction but can also be extended to any robot with external force compliance requirements.

In conclusion, this work provides an important breakthrough for the practical application of guide robots. We have proposed an effective local planner. Despite its effectiveness, the proposed framework has several limitations. It remains sensitive to perception delays and state estimation errors, and the computational load may limit deployment on low-power platforms. To address these issues, future work will explore lightweight approximations of MPC, tighter integration with vision-based dynamic obstacle and user tracking, and adaptive safety margins based on environmental uncertainty. In the future, we aim to integrate more generative Vision-and-Language Navigation (VLN) models to develop a map-less global path planner for guide robots, thereby enhancing the navigation capabilities of guide robots in previously unseen environments.

### Acknowledgment

This work was funded by the National Key Scientific Instrument and Equipment Development Project of NSFC Fig.(No. 51927809) and the National Natural Science Foundation of China (No.92248303).